\documentclass{spie}

\usepackage{graphicx}

\begin{document}

\title{Applying the Decisiveness and Robustness Metrics to Convolutional Neural Networks}
\author{Christopher A. George\supit{a}, Eduardo A. Barrera\supit{a}, and Kenric P. Nelson\supit{b}
\skiplinehalf
\supit{a}{Boston Fusion Corp, 70 Westview St Suite 100, Lexington MA, USA} \\
\supit{b}{Photrek, 50 Milk Street, Boston MA, USA}
}

\authorinfo{Further author information: Send correspondence to eduardo.barrera@bostonfusion.com}
\pagestyle{plain}
\setcounter{page}{1}
\maketitle

\begin{abstract}
We review three recently-proposed classifier quality metrics and consider their suitability for large-scale classification challenges such as applying convolutional neural networks to the 1000-class ImageNet dataset. These metrics, refered to as the ``geometric accuracy,'' ``decisiveness,'' and ``robustness,'' are based on the generalized mean ($\rho$ equals 0, 1, and -2/3, respectively) of the classifier's self-reported and measured probabilities of correct classification. We also propose some minor clarifications to standardize the metric definitions. With these updates, we show some examples of calculating the metrics using deep convolutional neural networks (AlexNet and DenseNet) acting on large datasets (the German Traffic Sign Recognition Benchmark and ImageNet). 
\end{abstract}


\pagenumbering{gobble}

\section{Introduction}
\label{sec:introduction}
Prediction accuracy, quantified as the number of correct predictions divided by the number of total predictions, has emerged as the \textit{de facto} measure of a classifier's quality. Modern deep-learning based classifiers, however, are well known for being both highly accurate and quite poor at predicting their own level of certainty. In particular, the ``logit'' values commonly used as a surrogate for probability have been shown to be vulnerable to a wide range of attacks that produce high-confidence but incorrect classification decisions \cite{fooled}. In light of this, it is worth considering a more general conception of accuracy.

In recent work \cite{nelson, nelson2}, Nelson argues that the \textit{geometric mean of the probabilities assigned to the correct classes}, rather than the prediction accuracy defined above, should be used as the key measure of a classifier's quality. He then generalizes this concept to introduce two additional metrics, which he calls \textit{decisiveness} (or decisive-biased accuracy) and \textit{robustness} (robust-biased accuracy) to provide further insight. Nelson shows that these metrics can be assessed using both the classifier's \textit{reported} probability and the classifier's \textit{measured} probability, giving insight into the model's degree of under- or over-confidence. 

In this work, we take up the challenge of applying these metrics to large-scale, deep-learning-based classifiers. By calculating the metrics for the classification decisions made by two convolutional neural networks applied to two real-world, imagery datasets, we give insight to both the metrics and the models. We also propose some minor clarifications to the original metric definition, allowing us to standardize these metrics. 

We organize the remainder of this paper as follows. In section \ref{nomenclature}, we define the terms we will use in the remainder of this discussion. We then review Nelson's past work in more detail (section \ref{review}) and then provide the full prescription for calculating our metrics (section \ref{proc}), highlighting clarifications to the original methodology. In section \ref{results}, we show the results of applying these metrics to our two datasets and two classifiers. We then conclude in section \ref{conclude}. 

\section{Nomenclature}
\label{nomenclature}
To avoid ambiguity, we define the following terms:
\begin{itemize}
\item \textbf{Prediction Accuracy}: the number of correct predictions divided by the number of total predictions. 
\item \textbf{Reported Probability}: the probability that a classifier assigns to a given class for a given target. These may be the ``logit'' values from a neural network, or may be calculated using any other technique. This can also be called ``model probability.''
\item \textbf{Measured Probability}: the actual probability that a given classification decision is correct, based on the classifier's historical performance on items with similar reported probabilities. Note, this could be generalized to consider other types of historical performance (e.g., per reported class), but we do not consider this here. This can also be called ``source probability.'' 
\item \textbf{Correct-Class Probability}: the probability (reported or measured) assigned to the correct class for a given target.
\item \textbf{Geometric, Arithmetic, or -2/3 Accuracy}: the geometric, arithmetic, -2/3, or generalized mean of the probabilities (where probabilities lower than $\gamma$ are set equal to $\gamma$ to avoid zero values). These quantities are defined for both the reported probabilities and the truth probabilities. 
\end{itemize}

\section{Review of Nelson's Past Work}
\label{review}
Machine-learning based classifiers are generally trained with the binary cross-entropy, also called the log loss. This binary cross-entropy represents the arithmetic mean of the log of the ``probability,'' $p(y_i)$, assigned to the correct class (measured state), $y_i$. In general, most neural-network based architectures define $p(y_i)$ as the logit value assigned to each class, after normalization with a softmax function. Mathematically, then, the cross-entropy is defined as:
\begin{equation} H = -\frac{1}{N} \sum_i  y_i \log(p(y_i))  \end{equation}
Optimizing this cross-entropy is mathematically equivalent to optimizing the \textit{geometric} mean of the probabilities. Nelson argues that this same quantity -- the geometric mean of the measured state probabilities (the classifier's reported probabilities of the correct classes) -- should be used so that the interpretation of performance is clear. We will call this quantity the \textit{geometric accuracy}. Nelson justifies this choice with arguments based on both Bayesian statistics and information theory, and cites a great body of literature going back to 1879 \cite{McAlister}. This approach becomes particularly important as advances in generalized entropy are incorporated.  In this case, a cumbersome body of proposed functions, all translate to use of the generalized mean of the probabilities as a spectrum of metrics.

The geometric mean, of course, is simply a special case ($\rho = 0$) of the generalized mean, which is defined as:
\begin{equation} M_\rho(x_1, \ldots, x_N) = \left( \frac{1}{N} \sum_{i=1}^N x_i^\rho \right)^{1/\rho} \end{equation}

Nelson's metrics are therefore as follows:
\begin{itemize}
\item \textbf{Geometric Accuracy}: the generalized mean of the correct-class probabilities with $\rho = 0$ (geometric mean). Mathematically, this represents the translation of the cross-entropy back to the probability domain, and so is a ``neutral-biased'' measure of the accuracy. 
\item \textbf{Decisiveness (Arithmetic Accuracy)}: the generalized mean of the probabilities with $\rho = 1$ (arithmetic mean). Mathematically, this would be equal to the prediction accuracy if the classifier perfectly reported its uncertainty. This arithmetic mean is therefore ``decisive-biased'' in that it is closely tied to the decision performance of the algorithm. Decisiveness is relatively insensitive to reported probabilities near zero, and therefore provides more sensitivity to the high-confidence region. 
\item \textbf{Robustness (-2/3 Accuracy)}: the generalized mean of the probabilities with $\rho = -2/3$ (-2/3 mean). Mathematically, the -2/3rds mean is the complement to the arithmetic mean because due to the conjugate relationship between positive and negative generalizations of the log-score \cite{nelson2}. This -2/3 mean is ``robustness-biased'' in that it is highly sensitive to reported probabilities near zero, and therefore provides more sensitivity to the low-confidence region. In particular, it assesses how well the classifier handles sources of severe error, such as events not included in the training. 
\end{itemize}
Together, decisiveness and robustness place bounds on the geometric accuracy. 

These metrics can be calculated both for the reported or measured probabilities (as defined in section \ref{nomenclature}). Indeed, Nelson proposes calculating both, and using their slope to determine underconfidence or overconfidence. In particular, the slope is defined by:
\begin{equation} s = \frac{d_{\textrm{truth}} - r_{\textrm{truth}}}{d_{\textrm{reported}} - r_{\textrm{reported}}} \end{equation}
where $d$ and $r$ represent decisiveness and robustness, respectively. Then, slopes greater than unity indicate underconfidence (as the measured confidence rises faster than the reported confidence), and slopes less than unity indicate overconfidence. 

We note also that both the geometric mean and the -2/3 mean are highly sensitive to low values (and, indeeed, a single zero value sets the entire metric to zero). This is partially by design (robustness in particular is designed to give sensitivity to the lower-probability tail), but we will avoid the extreme case of zero values by setting all probabilities below some threshold $\gamma$ to $\gamma$.  

\section{Metric Calculation Procedure and Clarifications}
\label{proc}

Calculating the reported metrics is simple: following Nelson's prescription, we calculate the generalized mean of the reported probability for each correct decision, setting all reported probabilities below some threshold $\gamma$ to $\gamma$.

Calculating the truth metrics is more complicated. Intuitively, to calculate a decision's ``measured probability'' we need to find several decisions with similar reported probabilities and calculate the fraction of correct decisions. We can then combine these measured probabilities with the generalized mean, and calculate the metrics as before. 

We implement this using histograms, largely following Nelson's original prescription. The prescription, however, specifies bins with ``equal amounts of data'' to ensure ``adequate data for the analysis.'' We must clarify this statement in three respects:
\begin{itemize}
\item \textbf{It is \textit{essential} that the bins have (approximately) equal data.} The overall metrics are an average of the metrics in each bin. If some bins have more data than others, the metrics will be distorted accordingly. Thus, even if evenly-spaced bins could provide adequate data, they would still not be an appropriate choice (unless each bin were weighted by its population). We allow for ``approximately'' equal data only to the extent that the number of data points may not be exactly divisible by the number of bins.
\item \textbf{By ``equal data bins,'' we mean that each bin should have (approximately) the same number of \textit{correct-class probabilities}.} The incorrect-class probabilities have no role in defining the bins, as the metrics by definition correspond to the correct-class probabilities. 
\item \textbf{We adjust our binning to avoid singularities (very high numbers of decisions with equal correct-class probabilities).} In particular, there is likely to be a singularity for correct-class probabilities equal to one. As needed, therefore, we exempt some bins from the equal-population requirement, but compensate by (a) limiting their width (for convenience, we assign overfull bins a width of $\gamma$) and (b) weighting overfull bins in proportion to their population when calculating the metrics. 
\end{itemize}

\noindent The final procedure for calculating the truth metrics is therefore as follows:
\begin{enumerate} 
\item Identify any singularities and assign such singularities a bin of width $\gamma$.
\item Bin the reported correct-class probabilities into $B$ equal-population bins (modulo divisibility issues).
\item Count the number of correct-class predictions $N_c$ and incorrect-class predictions $N_i$ in each bin.
\item Calculate the fraction correct for each bin, $f = N_c/(N_c + N_i)$.
\item Obtain the truth metrics by taking the generalized mean of these fractions, where we set any fractions below $\gamma$ equal to $\gamma$ and weight any overfull bins appropriately.
\end{enumerate}

\section{Numerical Experiments and Results}
\label{results}
We conduct numerical experiments by calculating these metrics for two convolutional neural network architectures and two datasets. Our neural networks are AlexNet \cite{AlexNet} and DenseNet \cite{DenseNet}, representing one of the earliest and most recent neural architectures, respectively. For the datasets, we considered the German Traffic Sign Recognition Benchmark (GTSRB) \cite{GTSRB} and ImageNet \cite{ImageNet}. The GTSRB has approximately 39K training images over 43 classes, while ImageNet has approximately 1.3M training images over 1000 classes. Both datasets have been widely used to benchmark computer vision architectures. We used $\gamma = 0.005$ for this work. 

Table \ref{tab:results} shows the results. 

\begin{table}[h]
\centering
\caption{Results. The slashes indicate reported / measured metrics, respectively}
\label{tab:results}
\vspace{0.05in}
\begin{tabular}{|c|c|c|c|c|}
\hline
 & AlexNet-GSSRB & DenseNet-GSSRB & AlexNet-ImageNet & DenseNet-ImageNet \\ 
\hline
Prediction Accuracy & 0.974 & 0.948 & 0.568 & 0.687 \\
\hline
Geometric Accuracy & 0.913 / 0.913 & 0.831 / 0.828 & 0.063 / 0.185 & 0.148 / 0.304 \\
\hline
Robustness & 0.603 / 0.679 & 0.437 / 0.505 & 0.054 / 0.054 & 0.085 / 0.087 \\
\hline
Decisiveness & 0.973 / 0.962 & 0.939 / 0.920 & 0.465 / 0.444 & 0.607 / 0.573 \\
\hline
\end{tabular}
\end{table}

\begin{figure*}[!h]
  \includegraphics[width=3.35in]{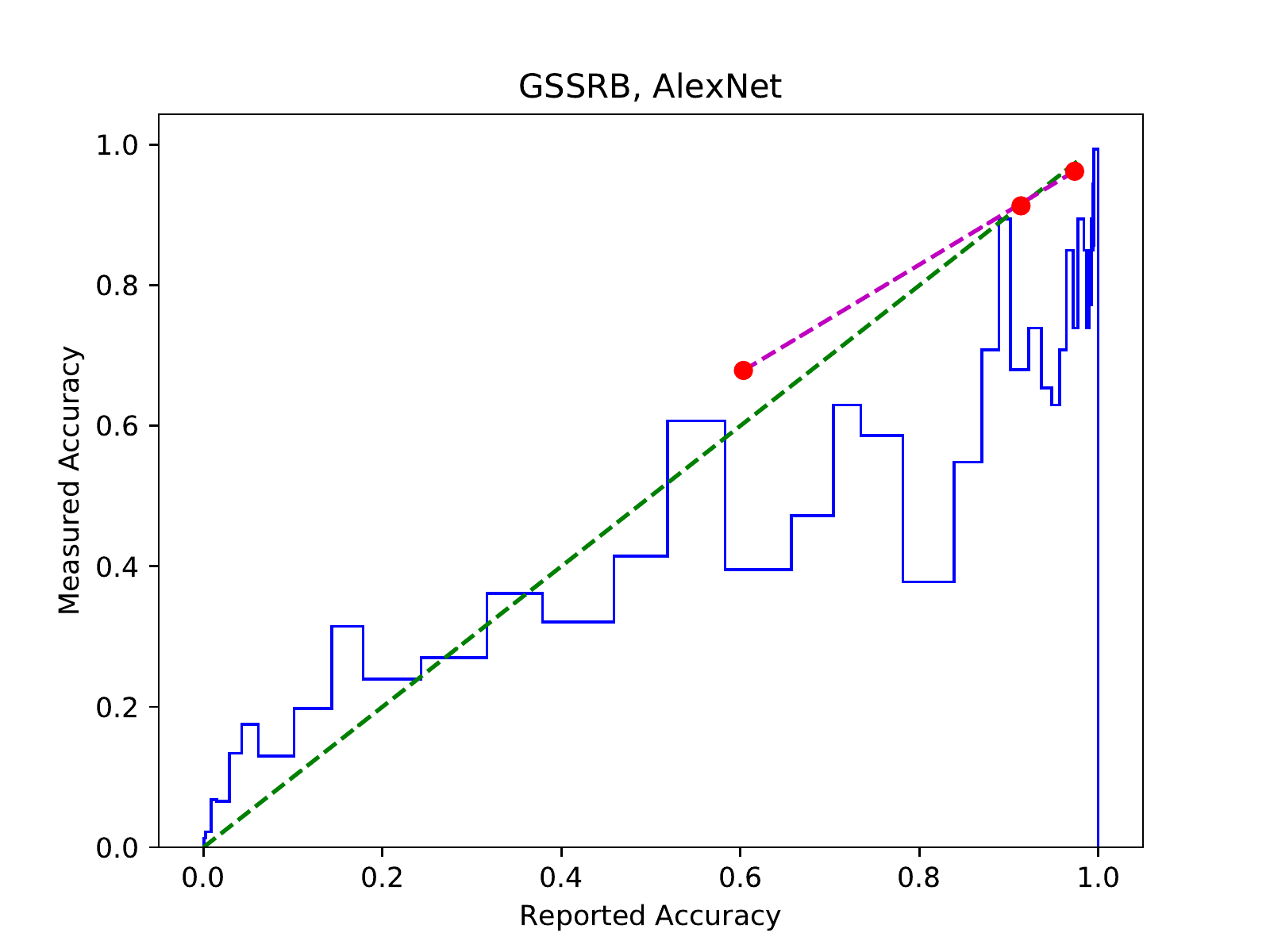}
  \includegraphics[width=3.35in]{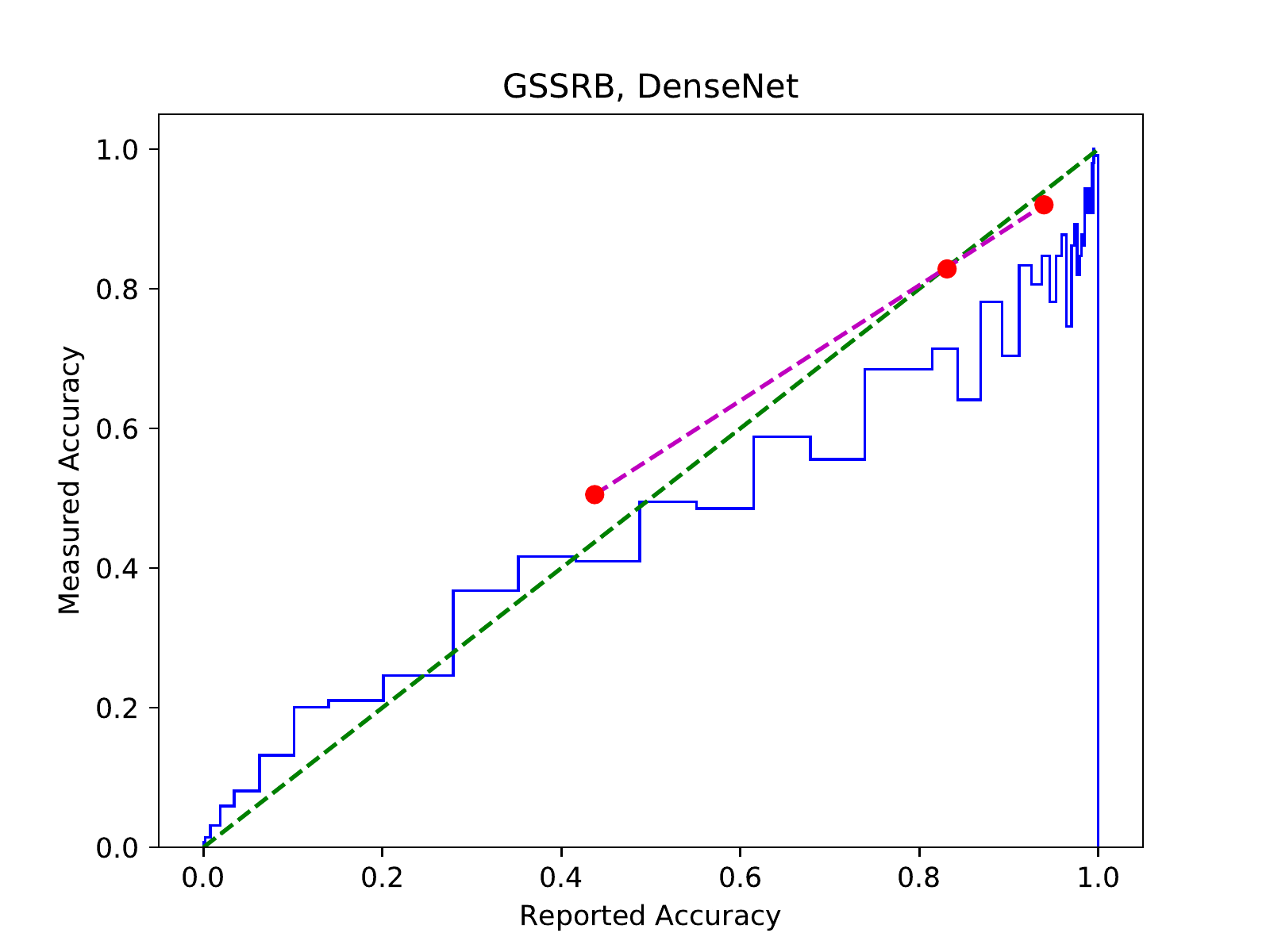}
  \includegraphics[width=3.35in]{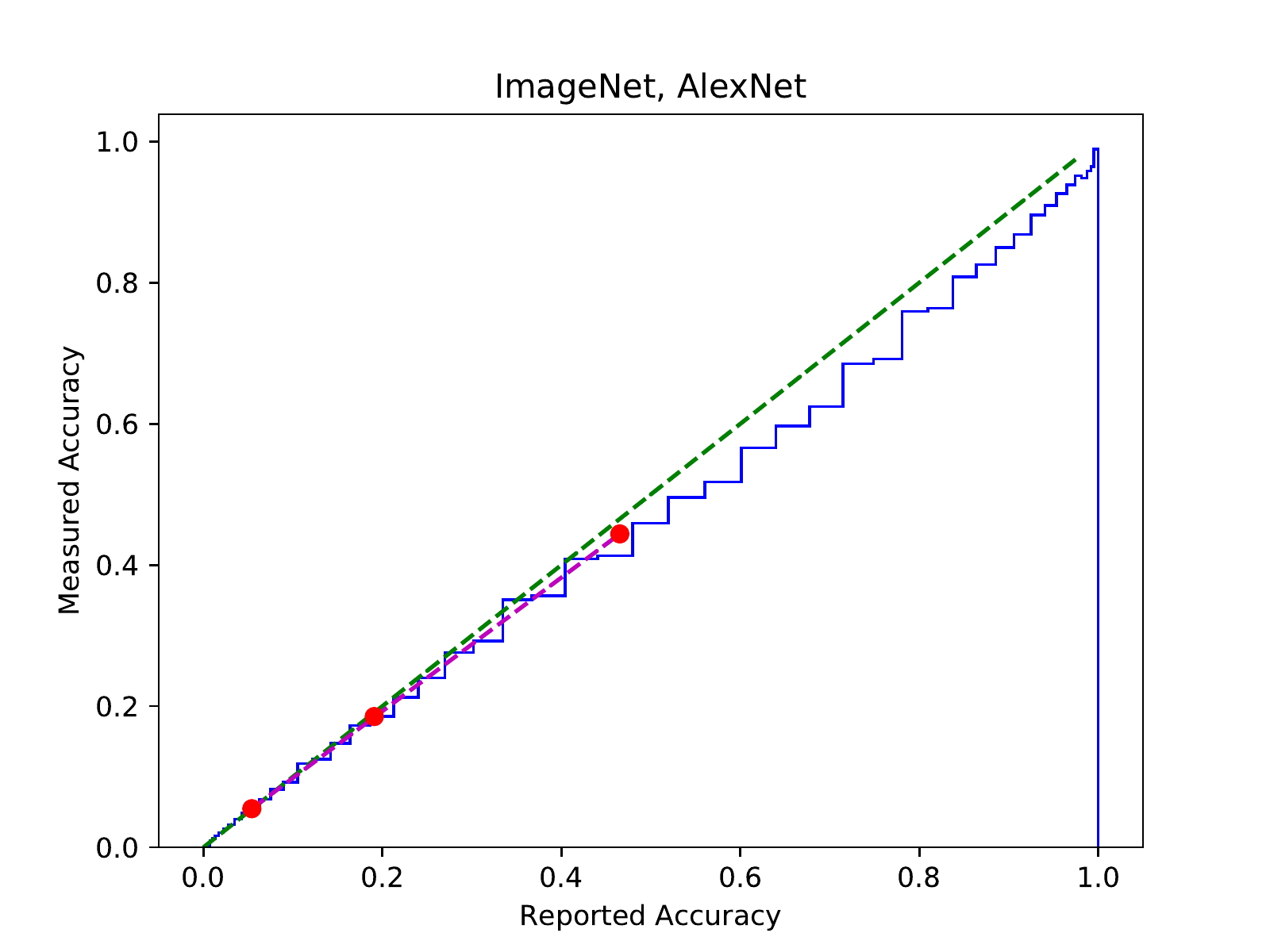}
  \includegraphics[width=3.35in]{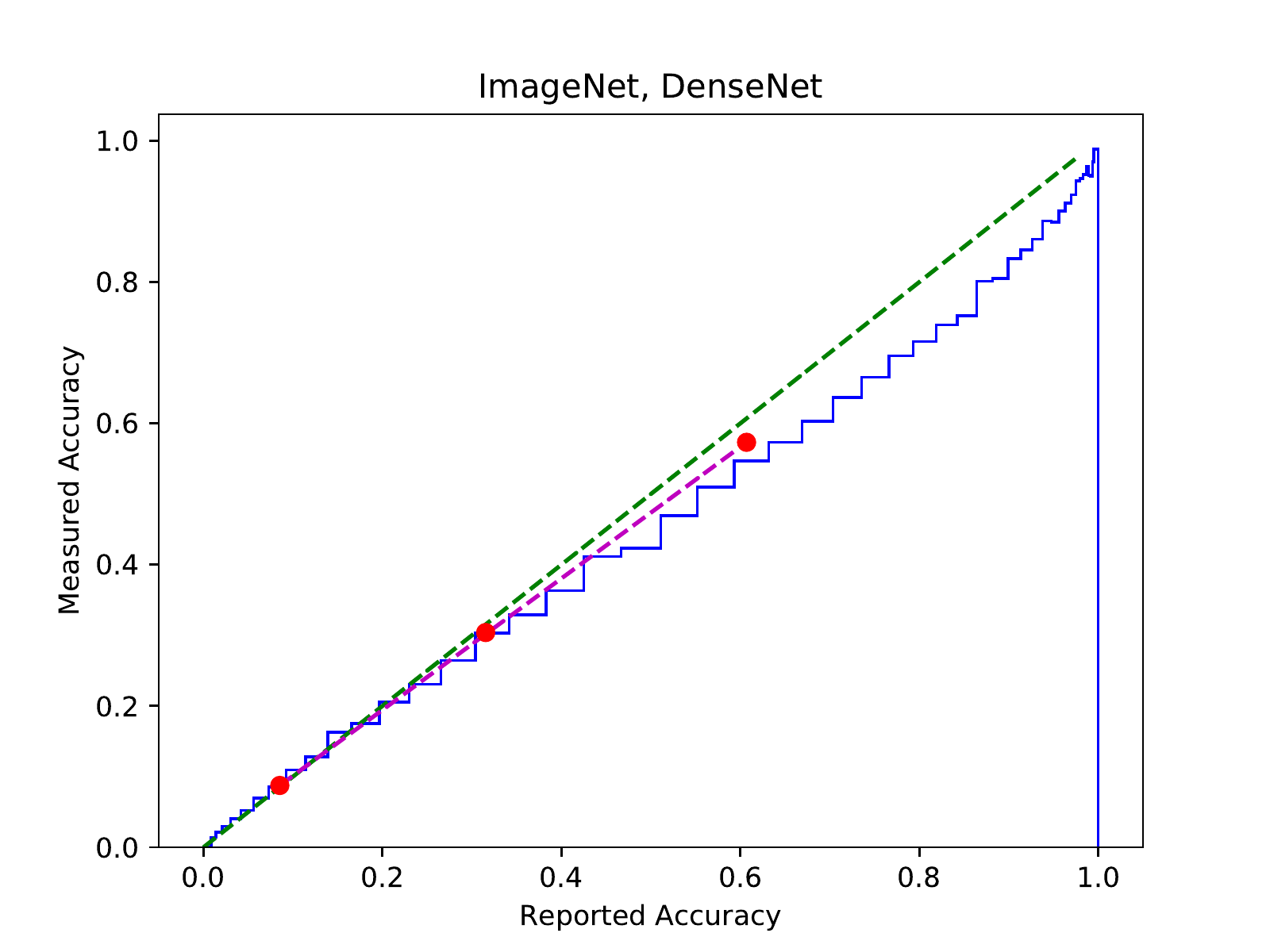}
  \centering
  \caption{Illustration of Reported Accuracy vs. Measured Accuracy for two datasets and two classifiers (as indicated). Note the variable-width bins. The green line reflects perfect alignment between reported and measured accuracy, while the red dots (from left to right) show the (reported, truth) values for the robustness, geometric accuracy, and decisiveness, respectively.}
  \label{fig:results}
\end{figure*}

Figure \ref{fig:results} shows a visualization of our results. We interpret this figure as follows:
\begin{itemize}
\item The blue histogram reprents the measured prediction accuracy ($f$) in each bin. Note the variable bin widths; since the classifiers are in general highly accurate, the bins are smallest on the right side of the graph (and the last bin includes all reported accuracies above 0.995). 
\item The green line has slope 1. A classifier that perfectly reports its uncertainty would be well-aligned with the green line. 
\item The red dots indicate the metrics as calculated with the reported probabilities (x-axis) and truth probabilities (y-axis). 
\item The magenta line shows the slope of the line connecting the decisiveness with the robustness. 
\end{itemize}

Since there is little difference between the reported and measured metrics, we conclude that the classifiers generally assess their probability correctly on the validation dataset. We find, however, that all four magenta lines in Figure \ref{fig:results} have slopes less than one; this implies that all four classifiers are slightly overconfident. We note also that the geometric accuracy and robustness are affected by probabilities near $\gamma$. Table \ref{tab:gamma} illustrates the effect of $\gamma$ on the metrics for one of our classifiers. 

\begin{table}[h]
\centering
\caption{Effect of $\gamma$ on the metrics for the DenseNet classifier on the GSSRB dataset}
\vspace{0.05in}
\label{tab:gamma}
\begin{tabular}{|c|c|c|c|}
\hline
$\gamma$ & robustness & geometric accuracy & decisiveness \\
\hline
.05 & .708 & .863 & .938 \\
\hline
.01 & .509 & .832 & .938 \\
\hline
.005 & .425 & .823 & .938 \\
\hline
.001 & .255 & .808 & .938 \\
\hline
0 & .019 & .790 & .938 \\
\hline
\end{tabular}
\end{table}

\section{Conclusion}
\label{conclude}
We have applied the generalized metrics, geometric accuracy, decisiveness, and robustness, to real-world, deep-learning-based classifiers on large-scale datasets. We have also clarified the binning scheme to define a well-defined procedure by which these metrics can be calculated across different classifiers. In particular, we found it necessary to set a minimum bound ($\gamma$) on all values when calculating the metrics; the value of $\gamma$ significantly affected geometric accuracy and (especially) robustness. 

We found that the source and model metrics were remarkably consistent. Integrating decisiveness and robustness into the classifier's loss function is a promising future direction, and has indeed already been partially considered \cite{nelson3}. 

\section*{Acknowledgment}
This work was supported by the Air Force Research Laboratory under contract number FA8750-17-C-0282. 

\bibliographystyle{unsrt}
\bibliography{LEARN}

\end{document}